\documentclass[letterpaper, 10 pt, conference]{ieeeconf}  

\usepackage{cite}
\usepackage{caption} 
\usepackage{graphicx}
\usepackage{subfigure} 
\usepackage{amsmath,amssymb,amsfonts}
\usepackage{algorithmic}
\usepackage{graphicx}
\usepackage{textcomp}
\usepackage{xcolor}
\usepackage{float}
\usepackage{placeins}
\usepackage{caption}
\usepackage{subcaption}
\usepackage{hyperref}
\newcommand{\method}{Offline RL-VLM-F}
\IEEEoverridecommandlockouts                              

\usepackage{booktabs}

\title{\LARGE \bf
Real-World Offline Reinforcement Learning from Vision Language Model Feedback
}

\author{Sreyas Venkataraman$^{*}$, Yufei Wang$^{*}$, Ziyu Wang, Navin Sriram Ravie, Zackory Erickson$^{\dagger}$, David Held$^{\dagger}$ 
\thanks{*Equal contribution. $^{\dagger}$ Equal advising.}
\thanks{
Sreyas Venkataraman is with Indian Institute of Technology, Kharagpur. Ziyu Wang is with IIIS, Tsinghua University. Yufei Wang, Zackory Erickson, and David Held is with the Robotics Institute, Carnegie Mellon University. This work was done when Sreyas Venkatarman and Ziyu Wang were visiting the Robotics Institute, Carnegie Mellon University. 
        {\tt\small \{sreyasv, yufeiw2, ziyuw2, nravie, zerickso, dheld\} @andrew.cmu.edu}}%
}

\begin{document}

\maketitle
\thispagestyle{empty}
\pagestyle{empty}

\begin{abstract}

Offline reinforcement learning can enable policy learning from pre-collected, sub-optimal datasets without online interactions. This makes it ideal for real-world robots and safety-critical scenarios, where collecting online data or expert demonstrations is slow, costly, and risky. However, most existing offline RL works assume the dataset is already labeled with the task rewards, a process that often requires significant human effort, especially when ground-truth states are hard to ascertain (e.g., in the real-world).
In this paper, we build on prior work, specifically RL-VLM-F, and propose a novel system that automatically generates reward labels for offline datasets using preference feedback from a vision-language model and a text description of the task. 
 Our method then learns a policy using offline RL with the reward-labeled dataset. We demonstrate the system's applicability to a complex real-world robot-assisted dressing task, where we first learn a reward function using a vision-language model 
on a sub-optimal offline dataset, and then 
we use the learned reward to employ Implicit Q learning to develop an effective dressing policy. Our method also performs well in simulation tasks involving 
the manipulation of rigid and deformable objects, and significantly outperforms baselines such as behavior cloning and inverse RL. In summary, we propose a new system that enables automatic reward labeling and policy learning from unlabeled, sub-optimal offline datasets. Videos can be found on our \href{https://offline-rlvlmf.github.io/}{project website}\footnote{https://offline-rlvlmf.github.io/}.
\end{abstract}

\begin{figure*}[t]
    \centering
    \includegraphics[width=.92\textwidth]{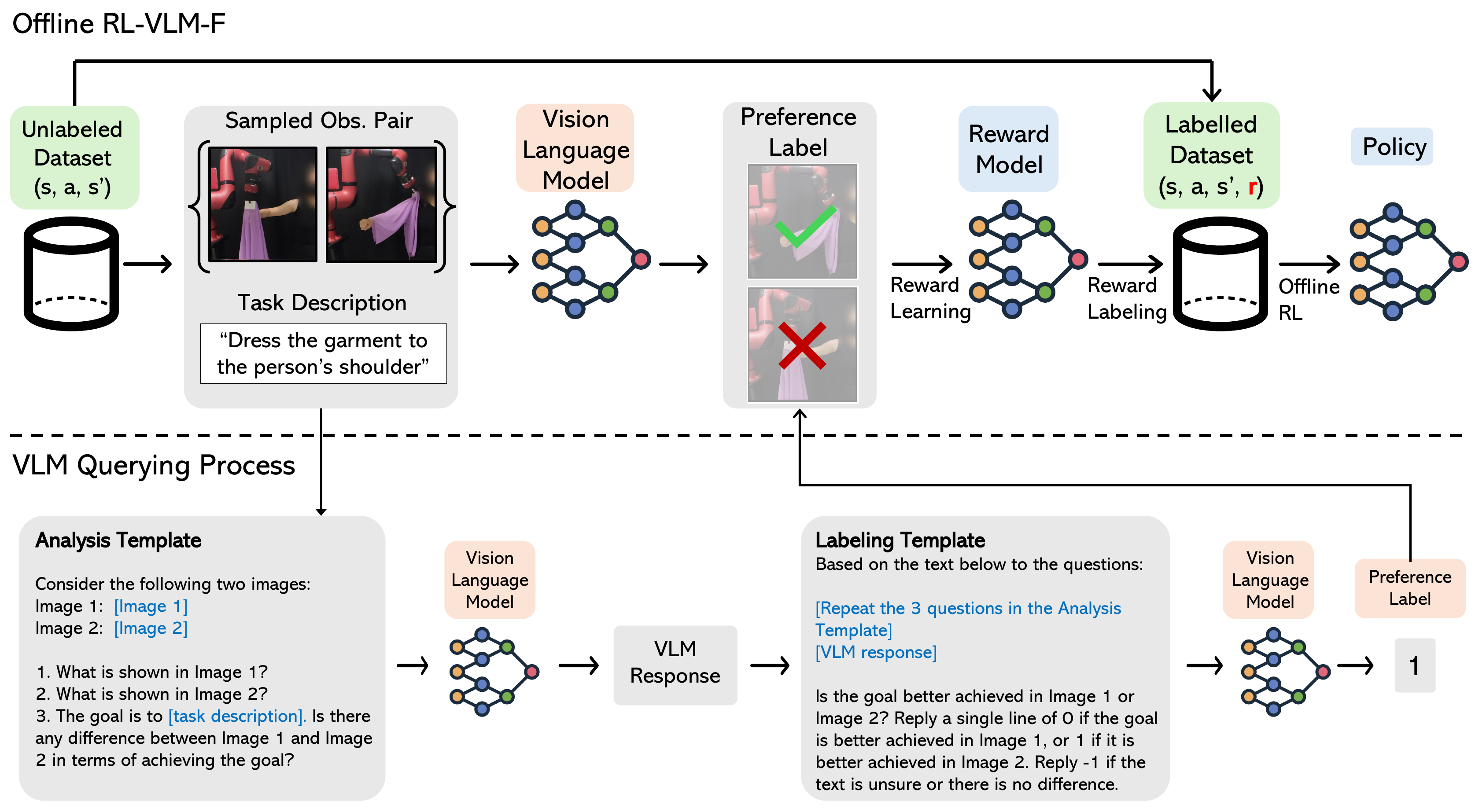}
    \caption{Top: Our system, \method{}, combines RL-VLM-F~\cite{wang2024rlvlmfreinforcementlearningvision} with offline RL for effective policy learning from unlabeled datasets. Given an unlabeled dataset without rewards, it first samples observation pairs and queries a Vision Language Model for preferences given a text description of the task. Using the preference labels, it then learns a reward model via preference-based reward learning. The learned reward model is utilized to label the entire offline dataset. Finally, it performs offline RL with the labeled dataset to learn a policy that solves the task. Bottom: We follow the same VLM querying process as in RL-VLM-F~\cite{wang2024rlvlmfreinforcementlearningvision}. It consists of two stages: the first is an analysis stage that asks the VLM to analyze and compare the two images; the second is the labeling stage, where the VLM generates the preference labels based on its own analysis from the first stage and the task description. }
    \label{fig:overview}
    \vspace{-0.2in}
\end{figure*}

\section{INTRODUCTION}

Offline reinforcement learning (RL) involves learning policies from a pre-collected, potentially sub-optimal offline dataset. As it does not require online interactions with the environment, it is particularly suited to scenarios where such interactions are impractical, e.g., learning policies with real robots. However, a key challenge remains: the need for the dataset to be labeled with the task rewards. 

Most prior work assumes that the offline dataset comes with labeled rewards and focuses on algorithm design~\cite{kostrikov2021offlinereinforcementlearningimplicit, kumar2020conservative, levine2020offline, fujimoto2021minimalist}. While reward labeling can be straightforward for simple tasks or those in simulation where the rewards can be generated based on low-level environment states, it becomes significantly more challenging for complex or real-world tasks where ground-truth states are not easily accessible. In such cases, manual reward labeling is often needed, making it a time-consuming bottleneck when applying offline RL algorithms. In this paper, we propose a system that automatically generates reward labels for a given offline dataset, enabling the learning of effective robot control policies from unlabeled, sub-optimal datasets. 

To automatically generate rewards, we build upon a prior method, Reinforcement Learning from Vision Language Model Feedback (RL-VLM-F)~\cite{wang2024rlvlmfreinforcementlearningvision}, which was originally designed and evaluated to automatically generate a reward function for a given task in the context of \emph{online} reinforcement learning. RL-VLM-F operates by querying a vision-language foundation model (VLM) to provide preference labels over pairs of the agent's image observations, based on a textual description of the task goal. The algorithm then learns a reward function from these preference labels. This approach has demonstrated effectiveness in online RL settings, where the agent interacts with the environment iteratively: gathering new interaction data, querying the VLM for labels, updating the reward function, and relabeling the agent's experiences with the updated reward. However, this method has not been tested in an offline setting, nor has it been applied to real-world robotics tasks.

In this paper, we adapt RL-VLM-F for the offline RL setting. 
Specifically, 
we query a VLM to generate a preference dataset from the given offline dataset. Subsequently, we learn a reward function from the generated preference dataset, and we use the reward function to label the given dataset.
The dataset with the labeled rewards can then be utilized with existing frameworks for offline reinforcement learning to learn a control policy. An overview of our system is shown in Figure~\ref{fig:overview}.
Furthermore, we demonstrate the effectiveness of the proposed system by applying it to a complex real-world robot-assisted dressing task, learning a point-cloud based reward and policy from a sub-optimal unlabeled dataset.

We chose to build our system based on RL-VLM-F due to its ability to learn rewards directly from visual observations without requiring access to low-level state information, making it particularly well-suited for real-world applications. Many other prior works have explored the use of foundation models, e.g., large language models (LLMs), as a substitute for human supervision in generating reward functions~\cite{xie2024text2rewardrewardshapinglanguage, wang2024robogen, ma2023eureka, yu2023language, klissarov2023motif}. However, most of these efforts have focused on the online RL setting, and express reward functions as code, necessitating access to the environment code and low-level ground-truth state information~\cite{xie2024text2rewardrewardshapinglanguage, wang2024robogen, ma2023eureka}. This reliance poses challenges in high-dimensional environments, such as deformable object manipulation, and for real-world tasks where low-level states are often inaccessible.
Some other works employ contrastively trained vision-language models (VLMs), such as CLIP and BLIP scores, as the reward signals. Yet, these scores tend to be noisy and exhibit high variance, limiting their effectiveness~\cite{sontakke2024roboclip,rocamonde2023vision,  ma2023liv}.



By combining RL-VLM-F with offline reinforcement learning, our system, named \textit{\method{}}, eliminates the need for ground-truth state information or environment code in reward generation, and does not assume dataset optimality. This flexibility enables policy learning from unlabeled, pre-collected sub-optimal datasets.
We evaluate \method{} in simulation across a variety of tasks, including classic control, rigid, articulated, and deformable object manipulation, using diverse datasets with varying levels of optimality. Our results demonstrate that our method can effectively learn policies to solve a wide range of tasks, outperforming existing baselines in Behavioral Cloning (BC) and inverse reinforcement learning (IRL). Additionally, we validated the effectiveness of our approach in a complex real-world robot-assisted dressing task, where our method learns an effective dressing policy from a sub-optimal real-world dataset, outperforming alternative baselines.
In summary, we make the following contributions in this paper:
\begin{itemize}
\setlength{\itemsep}{0pt}  
  \setlength{\leftskip}{-10pt} 
    \item We introduce a system that extends RL-VLM-F by utilizing vision-language models (VLMs) to automatically generate reward labels for unlabeled datasets in the context of offline reinforcement learning.
    \item We demonstrate the efficacy of our method in generating reward functions and learning policies across a range of simulation tasks, including classic control, rigid, articulated, and deformable object manipulation, with diverse datasets of different levels of optimality. Our method achieves better performance over baselines. 
    \item We apply our method to a complex real-world robot-assisted dressing task and show it can automatically learn a reward and a dressing policy from point cloud observations with a sub-optimal dataset, outperforming baselines.
\end{itemize}

\section{Related Work}
\textbf{Inverse Reinforcement Learning (IRL).} 
Similar to our system, IRL methods~\cite{IRL-1, algoIRL} seek to learn a reward function from expert demonstrations, such that the expert policy is optimal with the learned reward. IRL methods usually assume that the given dataset is optimal and perfectly solves the task. In contrast, our method does not assume the given dataset to be optimal when learning the reward, as it only requires a textual description of the task goal and access to visual observations of the task execution. One of the most classic IRL algorithms is Maximum Entropy IRL~\cite{ziebart2008maximum}. Many new algorithms have been proposed in recent years, such as those that leverage adversarial training, and are shown to achieve better performances~\cite{ho2016generative, finn2016guided, fu2017learning, ni2021f, finn2016connection}.

\textbf{Behavior Cloning (BC).} Behavior cloning~\cite{BC} is a classic imitation learning method where an agent learns to perform tasks by mimicking expert demonstrations. The objective in BC is to learn a policy that maps states to actions, often achieved via supervised learning by minimizing the discrepancy between the agent's and the expert's actions. BC methods do not require the dataset to contain the reward information, but they usually assume the demonstrations are optimal in solving the task.
Many recent works have shown that Behavior Cloning can be a strong method in learning complex robotic manipulation policies with the right policy representation~\cite{florence2022implicit, chi2023diffusion, ze20243d, seita2023toolflownet, shafiullah2022behavior  }. 
Our method uses offline reinforcement learning with the learned reward and does not require either labeled rewards or demonstrations to be optimal. 

\textbf{Reward Generation from foundation models. } Designing the right reward function, also known as reward engineering, has always been a challenge for reinforcement learning~\cite{gupta2022unpacking, henderson2018deep}. Recently, many works have looked at using foundation models, such as large language models or vision language models, to automatically design reward functions for a given task~\cite{yu2023language, xie2024text2rewardrewardshapinglanguage, wang2024robogen, ma2023eureka, sontakke2024roboclip, rocamonde2023vision, klissarov2023motif, wang2024rlvlmfreinforcementlearningvision}. Almost all of these prior works focus on generating the reward in the online reinforcement learning setting; in contrast, we focus on the offline RL setting. None of these prior works have shown they are able to generate a reward function directly in the real world, as most of them require accessing the environment or ground-truth low-level states~\cite{ma2023eureka, wang2023one, yu2023language, xie2024text2rewardrewardshapinglanguage, klissarov2023motif}. We are the first to show that RL-VLM-F~\cite{wang2024rlvlmfreinforcementlearningvision}, the prior work that we build on, can be extended to directly generate effective rewards from sub-optimal datasets in the real world.

\section{Preliminaries}


We consider a standard discounted Markov Decision Process (MDP) formulation of reinforcement learning\cite{Sutton1998}, defined by the tuple $(\mathcal{S},\mathcal{A}, P, \mathcal{R}, \gamma)$, where $\mathcal{S}$ is the state space; $\mathcal{A}$ is the action space; $\mathcal{R}$ is the set of possible rewards;
$\gamma \in$ [0,1] is the discount factor; and $\displaystyle P :S \times \mathcal{A} \times \mathcal{S} \rightarrow$ [0,1] is the state transition probability function. For a given state $\displaystyle s \in \mathcal{S}$ and action $\displaystyle a \in \mathcal{ A}$, the agent transitions to state $s'$ gaining a reward $r \in \displaystyle\mathcal{R}$ with the probability $\displaystyle P(s', r|s, a)$.
The Q function $Q_{\pi}(s, a)$ of policy $\pi$ is defined as the sum of future discounted rewards starting from state $s$, taking action $a$, and following policy $\pi$. The goal is to learn an optimal policy $\pi$ that maximizes the expected reward over time.


\textbf{Preference-based Reinforcement Learning.}
Our work builds upon preference-based RL, in which a reward function is learned from preference labels over the agent's behaviors \cite{christiano2023deepreinforcementlearninghuman,lee2021pebblefeedbackefficientinteractivereinforcement}. 
Formally, a segment $\sigma$ is a sequence of states $\{s_1, \ldots, s_H\}$, where $H \geq 1$. In this work, we consider the case where the segment is represented using a single image. Given a pair of segments $(\sigma_0, \sigma_1)$, an annotator gives a feedback label $y$ indicating which segment is preferred: $y \in \{-1, 0, 1\}$, where 0 indicates the first segment $\sigma_0$ is preferred, 1 indicates the second segment $\sigma_1$ is preferred, and -1 indicates they are incomparable or equally preferable.
Given a parameterized reward function $r_{\psi}$ over the states, we follow the standard Bradley-Terry model \cite{Bradley1952RankAO} to compute the preference probability of a pair of segments:
\begin{equation}
P_{\psi} [\sigma_1 \succ \sigma_0] = \frac{\exp \left( \sum_{t=1}^{H} r_{\psi} (s_{1, t}) \right)}{\sum_{i \in \{0,1\}} \exp \left( \sum_{t=1}^{H} r_{\psi} (s_{i, t}) \right)},
\label{eq:bt_model}
\end{equation}
where $\sigma_i \succ \sigma_j$ denotes segment $i$ is preferred to segment $j$. Given a dataset of preferences $D = \{(\sigma_0^i, \sigma_1^i, y_i)\}$, preference-based RL algorithms optimize the reward function $r_{\psi}$ by minimizing the following loss:
\begin{align}
L_{\text{Reward}} = &- \mathbb{E}_{(\sigma_0, \sigma_1, y) \sim D} \left[ \mathbb{I}\{y = (\sigma_0 \succ \sigma_1)\} \log P_{\psi} [\sigma_0 \succ \sigma_1] \right. \nonumber \\
&+ \left. \mathbb{I}\{y = (\sigma_1 \succ \sigma_0)\} \log P_{\psi} [\sigma_1 \succ \sigma_0] \right]
\label{eq:reward_loss}
\end{align}


\textbf{Offline RL and Implicit Q Learning (IQL)~\cite{kostrikov2021offlinereinforcementlearningimplicit}}. We consider the standard offline RL setting, in which the agent has no access to the online environment interactions but instead learns from a fixed dataset of transitions $D = \{s_i, a_i, r_{\psi}(s_i, a_i), s'_i\}_{i=1}^N$ of size $N$. These transitions can be collected with a behavior policy $\pi_{\beta}$, which might be a mixture of several sub-optimal policies. In contrast to prior work that assumes the reward is given, here we assume the reward is labeled by a learned function $r_\psi$, which in our case is learned using preference labels provided by a VLM (more details in Section~\ref{sec:system}). 

For all our experiments, we adopt Implicit Q-Learning (IQL)\cite{kostrikov2021offlinereinforcementlearningimplicit} as the base offline RL algorithm.
IQL predicts an upper expectile of the TD-target, which approximates the maximum of $r(s, a) + \gamma Q_\theta(s', a')$ over actions $a'$ constrained to the dataset actions. 
IQL trains a separate value and Q function. The value function $V_\psi$ is trained to approximate an expectile purely with respect to the action distribution:
\begin{equation}
L_V(\psi) = \mathbb{E}_{(s,a) \sim D} \left[L_2^\tau (Q_{\hat{\theta}}(s, a) - V_\psi(s))\right],
\end{equation}
where $L_2^\tau(u) = |\tau - \mathbf{1}(u < 0)| u^2$.
This value function is then used to train the Q-function: 
\begin{equation}
L_Q(\theta) = \mathbb{E}_{(s,a,s') \sim D} \left[(r(s,a) + \gamma V_\psi(s') - Q_\theta(s,a))^2\right].
\end{equation}
The policy $\pi$ is extracted from the Q function via advantage weighted regression (AWR)~\cite{peng2019advantage}:
\begin{equation}
L(\phi) = \mathbb{E}_{s,a \sim D} \left[ \exp(\beta(Q_\theta(s, a) - V_\psi(s))) \log \pi_\phi(a \mid s) \right],
\end{equation}
where $\beta$ denotes an inverse temperature.

\section{System}
\label{sec:system}
Our system, \method{}, consists of two phases: the reward labeling phase, which is built on top of RL-VLM-F~\cite{wang2024rlvlmfreinforcementlearningvision}, and the policy learning phase, based on Implicit Q Learning~\cite{kostrikov2021offlinereinforcementlearningimplicit}. Figure~\ref{fig:overview} provides an overview of the system. 
We describe each of the two phases in detail below. 

\subsection{Reward labeling of the offline dataset}
In the reward labeling phase, we aim to learn a reward model to label all of the transitions in the offline dataset. We employ RL-VLM-F~\cite{wang2024rlvlmfreinforcementlearningvision} to perform this labeling. We assume access to a text description of the task and that the offline dataset contains visual observations, e.g., images of the states. The labeling process has the following three steps.

\textbf{Sample Observations:} We begin by randomly sampling pairs of image observations from the offline dataset. The sampled image observation pairs, together with the text description of the task goal, are input to the VLM. 

\textbf{Query VLM:} We query the VLM to obtain preferences over the sampled observation pairs. The VLM evaluates the pair based on the provided task description and image observations. The output preference from the VLM is stored as labels. 
    As shown in the bottom part of  Fig.~\ref{fig:overview}, we follow the procedure from RL-VLM-F to query the VLM. The querying process involves two stages: analysis and labeling. In the analysis stage, we query the VLM for detailed responses describing and comparing how well each of two images achieves the task goal.  In the labeling stage, we use the VLM-generated text responses to extract a preference label between the images. Specifically, we ask the VLM to generate a preference label $y \in \{-1, 0, 1\}$, where 0 indicates the first image is better, 1 indicates the second image is better, and -1 indicates no discernible differences, based on its own response from the analysis stage. Image pairs labeled as -1 are not used to train the reward model. For both stages, we use a fixed prompt template and only fill in different text descriptions for different tasks.
    See Fig.~\ref{fig:overview} for the prompt templates we used for these two stages.

\textbf{Preference-based reward learning:} Using the stored preference labels, we follow the Bradley-Terry model as in Eq.~\ref{eq:bt_model} and learn a reward model using the loss in Eq.~\ref{eq:reward_loss}. The reward model is trained until it converges on the entire set of stored preference labels.

\subsection{Policy Learning from the labeled dataset}
In the policy learning phase, we first label the entire offline dataset using the learned reward model. From the labeled dataset, we then learn a policy using Implicit Q Learning.

\subsection{Implementation Details}
In the real world robot-assisted upper-body dressing task, we do not have access to the low-level states of the task for two reasons: 1) the cloth is highly deformable and does not have a known compact low-level state; 2) even if we manually define a low-level state as e.g., key points on the cloth, it is challenging to accurately estimate these key points reliably from sensory observations. Therefore, we choose to just represent the cloth, along with the human arm, using point clouds obtained from a depth camera, as was done in prior work~\cite{wang2023one}. We use this point cloud representation for both learning the reward model and the policy. Note that when querying the VLM, we still provide it with pairs of image observations, because the VLM was trained from images; 
however; after we store the preference labels from the VLM, we can learn the reward model using the corresponding point cloud observations. In simulation, we learn all the reward models from images, and the policy with low-level ground-truth states provided by the simulator. 
For the real-world dressing task, both the reward model and the policy are represented using the PointNet++~\cite{qi2017pointnet++} architecture; for all simulation tasks, the reward model is a ResNet model~\cite{he2016deep} and the policy is an MLP. All our experiments were conducted on 1 Nvidia RTX 3090/4090.

\begin{figure}[t]
    \centering
    \includegraphics[width=\columnwidth]{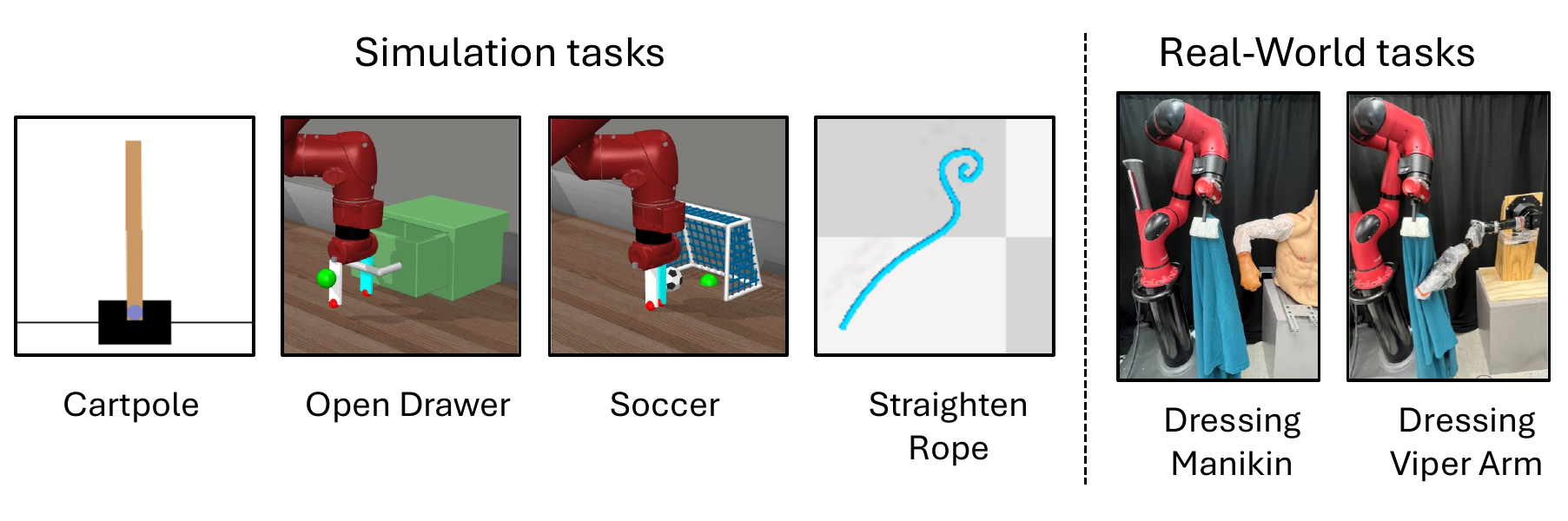}
    \vspace{-0.3in}
    \caption{The tasks that we evaluate our method on. The left four images visualize the simulation tasks. The rightmost two images show the real-world dressing task setup: the first shows the manikin that is being dressed; the second shows the ViperX 300 S arm that is being dressed. }
    \label{fig:tasks}
    \vspace{-0.2in}
\end{figure}

\begin{table*}[t]
\centering
\scriptsize
\caption{Performance of all compared algorithms in 4 simulation tasks. For each task, we test with 3 datasets of different levels of optimalities. For Open Drawer and Soccer, the number represents the success rate; for other tasks the number represents the ground-truth return. The best performing method other than IQL-GT Reward is bolded. }

\begin{tabular}{llcccccccc}
\toprule
Task            & Dataset & \begin{tabular}[c]{@{}c@{}}IQL\\GT reward \end{tabular}        & \begin{tabular}[c]{@{}c@{}}Offline\\RL-VLM-F (Ours) \end{tabular}    & \begin{tabular}[c]{@{}c@{}}IQL\\Avg. Reward\end{tabular}    & BC & Diffusion Policy  & GAIL       & CLIP Reward \\ \midrule
                & Random  & 0.99 (0.02)          & \textbf{0.91 (0.08)}   & 0.19 (0.12)           & 0.04 (0.06)                 & 0.08 (0.06)     & 0.03 (0.05) & 0.00 (0.00) \\ 
\begin{tabular}[c]{@{}c@{}}Open\\Drawer\end{tabular}    & Medium  & 0.95 (0.06)          & {0.85 (0.13)}         & 0.84 (0.14)           & \textbf{0.88 (0.10)}        & 0.00 (0.01)     & 0.00 (0.00) & 0.24 (0.11) \\ 
                & Expert  & 0.99 (0.01)          & {0.99 (0.01)}         & \textbf{1.00 (0.00)}   & \textbf{1.00 (0.00)}        & \textbf{1.00 (0.00)}  & 0.07 (0.09) & 0.99 (0.00) \\ \midrule

                & Random  & 0.30 (0.15)          & \textbf{0.23 (0.11)}   & 0.11 (0.09)           & 0.11 (0.08)                 & 0.13 (0.14)     & 0.01 (0.02) & 0.15 (0.01) \\ 
Soccer          & Medium  & 0.19 (0.10)          & 0.15 (0.10)           & 0.19 (0.10)           & \textbf{0.20 (0.12)}        & 0.10 (0.12)     & 0.00 (0.01) & 0.16 (0.02) \\ 
                & Expert  & 0.42 (0.13)          & 0.42 (0.12)           & 0.41 (0.13)           & 0.41 (0.13)                 & \textbf{0.49 (0.17)}  & 0.04 (0.06) & 0.28 (0.06) \\ \midrule

                & Random  & -93.54 (20.22)       & \textbf{-212.87 (36.84)} & -2537.62 (94.90)      & -1816.98 (109.75)           & -2063.65 (89.31) & -1683.31 (N/A) & -2672.38 (3.07) \\ 
Cartpole        & Medium  & -98.85 (49.86)       & \textbf{-144.56 (75.11)} & -236.77 (27.70)       & -165.16 (14.03)             & -1875.05 (149.51) & -431.31 (N/A) & -270.32 (9.28) \\ 
                & Expert  & -83.93 (14.89)       & \textbf{-78.77 (0.83)}   & -81.46 (5.18)         & -79.19 (0.99)               & -1423.32 (309.85) & -433.64 (N/A) & -79.69 (0.23) \\ \midrule

                & Random  & 20.93 (0.28)         & \textbf{16.70 (1.12)}   & 8.71 (1.43)           & 12.73 (1.17)                & 12.01 (3.44)    & 11.75 (0.30) & 13.66 (0.75) \\ 
\begin{tabular}[c]{@{}c@{}}Straighten\\Rope\end{tabular} & Medium  & 14.34 (1.00)         & 14.56 (0.96)          & 14.15 (1.11)          & 14.17 (0.98)                & \textbf{20.27 (1.44)}  & 19.88 (1.53) & 14.17 (0.21) \\ 
                & Expert  & 20.58 (0.30)         & \textbf{20.54 (0.32)}   & 20.46 (30.83)         & \textbf{20.54 (0.31)}        & 20.27 (0.30)    & 12.33 (2.86) & 20.49 (0.08) \\ \bottomrule
\end{tabular}
\label{tab:sim_results}
\vspace{-0.2in}
\end{table*}

\section{Experiments}
\subsection{Simulation Experiments}
\subsubsection{Experiment setting}
We first evaluate our method on the following four simulation environments. 
\textit{1. Cartpole}~\cite{cartpole}: a classic control task where the goal is to balance a pole on a moving cart. 
\textit{2. Open Drawer}: an articulated object manipulation task from MetaWorld~\cite{yu2021metaworldbenchmarkevaluationmultitask} where a Sawyer robot needs to pull out a drawer.
\textit{3. Soccer}: A dynamic task from MetaWorld where a Sawyer robot must push a soccer ball into the goal. 
\textit{4. Straighten Rope}: a deformable object manipulation task from SoftGym~\cite{lin2021softgymbenchmarkingdeepreinforcement} where the goal is to straighten a rope from a random configuration. See Fig.~\ref{fig:tasks} for an illustration of these tasks. 

For each of the environments, we test our method with three different kinds of dataset optimality: 
    \textit{1. Random}: the transitions in the dataset are collected from a policy that just performs random actions;
    \textit{2. Medium}: the transitions in the dataset are collected from a partially trained RL policy with the ground-truth reward. Therefore, the dataset is of medium quality. 
    \textit{3. Expert}: the transitions in the dataset are collected from an RL policy trained until convergence with the ground-truth task reward. The trained RL policy can solve the task; therefore this dataset is considered as the expert dataset.
We note that there are no reward labels provided in any of these datasets. 

We compare our method with the following baselines:
\textit{1. Simple behavior cloning (BC)}: we perform standard behavior cloning on the dataset to learn the policy. The policy is represented as a simple MLP.  
    \textit{2. GAIL~\cite{ho2016generative}}: an inverse RL algorithm that uses adversarial training to learn a cost and policy from the given demonstrations.
    \textit{3. Diffusion Policy~\cite{chi2023diffusion}}: a state-of-the-art imitation learning algorithm that represents the policy as a denoising diffusion process.
     \textit{4. IQL with average reward}: this baseline labels all transitions in the dataset with the same reward, which is the average of the ground-truth reward of the dataset. It then performs IQL with the labelled dataset. Such a baseline has been proposed and used in prior work~\cite{shin2023benchmarks}. 
     \textit{5. CLIP reward~\cite{rocamonde2023vision}}, which computes the reward as the cosine similarity score between the CLIP~\cite{radford2021learning} embedding of the task description and the image observation. 
    \textit{6. IQL with ground-truth reward}: this baseline performs IQL on the dataset, using ground-truth rewards. This baseline is an oracle and provides an upper bound for the performance.
Among these baselines, Simple BC, GAIL, and diffusion policy usually assume the dataset to be optimal for learning effective policies. 

For IQL, we use the default parameters as used for D4RL benchmarking~\cite{fu2020d4rl}. The Q function is approximated by a twin Q Network, and a Gaussian Policy is used for the actor. For all methods, the policy learns with state observations. 
The evaluation metric is the ground-truth return for CartPole, RopeFlatten and success rate for Soccer and Drawer Open. We run each method with 3 seeds and report the mean and standard deviation. 
When querying the VLM to obtain preference labels for learning the reward, we use the cached preference labels provided by RL-VLM-F, which are generated using Gemini-1.0-Pro.

\begin{figure}[t]
    \centering
    \includegraphics[width=.9\linewidth]{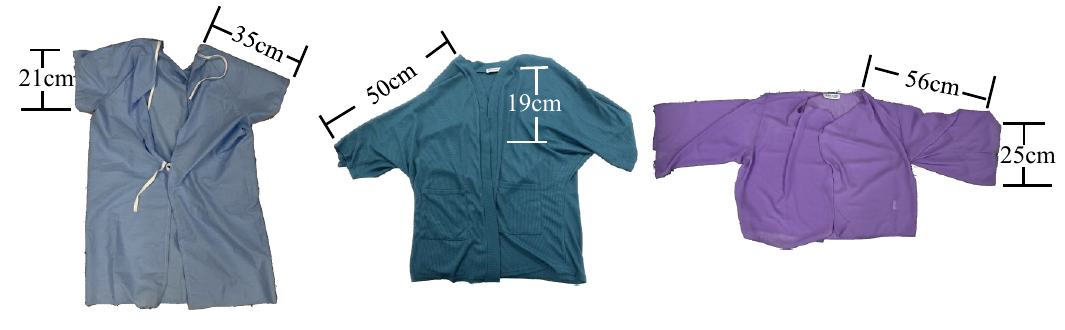}
    \vspace{-0.1in}
    \caption{We test 3 garments for the real-world dressing task: a hospital gown, a green jacket, and a purple jacket. }
    \label{fig:garment}
    \vspace{-0.2in}
\end{figure}

\subsubsection{Simulation Results}
The results for all simulation experiments can be found in Table~\ref{tab:sim_results}. As shown, for all tasks, when using the random dataset, our method outperforms all compared methods by a large margin (we do not consider the comparison to  IQL-GT Reward, which uses the ground-truth reward). This demonstrates the effectiveness of the learned reward using VLM preferences and that our system can produce effective policies from low-quality unlabeled datasets. 
When using the medium dataset, we find our method performs the best on Cartpole, and slightly worse than the best-performing methods on Open Drawer and Soccer. 
We also find it interesting that our method often performs better on the Random dataset than on the medium dataset; we hypothesize the reason to be that the medium dataset contains less exploration trajectories and thus less state coverage. 
As expected, the performance of most methods is best when trained on the expert dataset. 
We find Diffusion Policy to fail drastically for Cartpole even with expert demonstrations. We speculate the reason is that Cartpole is more dynamic, so the velocity information is more important.
Finally, we find the learning of GAIL to be unstable, and it fails to learn meaningful policies on most tasks, which has been reported in prior work as well~\cite{ni2021f}. The N/A in GAIL for Cartpole means two seeds run into NaN issues and the learning completely failed. 

\subsubsection{Ablation Study}
VLMs have errors when giving the preference label. We perform ablation studies to examine how robust our system is to errors in the preference labels. Specifically, we perform IQL with preference label obtained using the ground-truth reward, but with probability $p$ we flip the label of the pair. 
We test with $p=0, 0.25, 0.5, 0.75$. 
In addition, prior work~\cite{wang2024rlvlmfreinforcementlearningvision} has found that the error rate of the VLM increases as the underlying states of the two images are closer to each other. 
Therefore, we also test the case where the probability $p$ of flipping the label is inversely proportional to how close the two states are in the pair, termed "proximity flipping". 
The results are summarized in Table~\ref{tab:ablation}.
As shown, our system is robust to $25\%$ of the error in the preference labels. As the error rate $p$ increases, the performance decreases. 
As expected, given an error rate $p$, the performance is worse on dataset with lower quality (e.g., the Random dataset).
The proximity flipping results are slightly worse than that of having an error rate of 50\% but better than $75\%$. 

\begin{table*}
\centering
\caption{We study the performance of our method across a varying quality of preference datasets. Columns 0.25, 0.5, 0.75 represent the probability of the label being incorrect. The proximity-based flip is when the probability of the label being incorrect is proportional to how close the states are.}
\begin{tabular}{llccccc}
\toprule
Task            & Dataset & IQL-GT Reward & 0.25               & 0.5                & 0.75               & Proximity based Flip \\ \midrule

                & Random  & 0.99 (0.02)     & \textbf{0.99 (0.02)} & 0.51 (0.10)        & 0.00 (0.00)        & 0.00 (0.01) \\ 
Open Drawer     & Medium  & 0.95 (0.06)     & \textbf{0.52 (0.17)} & 0.16 (0.13)        & 0.01 (0.00)        & 0.02 (0.01) \\
                & Expert  & 0.99 (0.01)     & \textbf{0.99 (0.00)} & \textbf{0.99 (0.00)} & 0.92 (0.04)        & 0.96 (0.05) \\ \midrule

                & Random  & 0.30 (0.15)     & 0.15 (0.01)        & 0.15 (0.01)        & 0.15 (0.01)        & \textbf{0.19 (0.04)} \\
Soccer          & Medium  & 0.19 (0.10)     & 0.20 (0.04)        & 0.17 (0.03)        & 0.15 (0.03)        & \textbf{0.21 (0.03)} \\
                & Expert  & 0.42 (0.13)     & 0.26 (0.06)        & \textbf{0.30 (0.08)} & 0.28 (0.01)        & 0.28 (0.05) \\ \midrule

                & Random  & -93.54 (20.22)  & \textbf{-89.72 (0.77)} & -2677.65 (1.86)     & -2381.72 (24.71)     & -214.92 (8.19) \\
CartPole        & Medium  & -98.85 (49.86)  & \textbf{-99.16 (5.31)} & -259.92 (6.35)      & -773.81 (135.11)     & -650.91 (147.03) \\
                & Expert  & -83.93 (14.89)  & \textbf{-78.61 (0.10)} & -79.72 (0.24)       & -117.26 (28.18)     & -85.69 (11.48) \\ \midrule

                & Random  & 20.93 (0.28)    & \textbf{18.07 (0.39)} & 14.63 (1.26)        & 9.19 (0.45)         & 17.61 (0.31) \\
Straighten Rope & Medium  & 14.34 (1.00)    & 14.33 (0.28)        & \textbf{14.37 (0.16)} & 14.27 (0.28)        & 14.05 (0.16) \\
                 & Expert  & 20.58 (0.30)    & \textbf{20.54 (0.04)} & 20.41 (0.04)        & 20.50 (0.07)        & 20.51 (0.03) \\ \bottomrule
\end{tabular}
\label{tab:ablation}
\vspace{-0.2in}
\end{table*}


\begin{figure}[t]
    \centering
    \includegraphics[width=.8\linewidth]{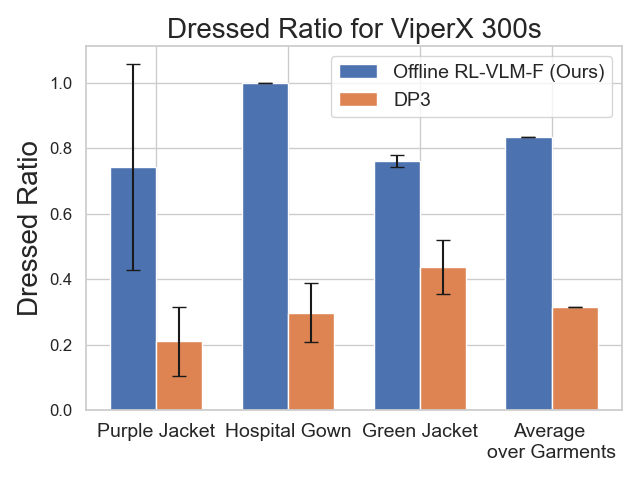}
    \vspace{-0.1in}
    \caption{Dressed ratio of our method and DP3 on the ViperX 300 S arm. As shown, our method achieves higher dressed ratios on all three garments. }
    \vspace{-0.2in}
    \label{fig:real-world-viper}
\end{figure}

\begin{figure}[t]
    \centering
    \includegraphics[width=.8\linewidth]{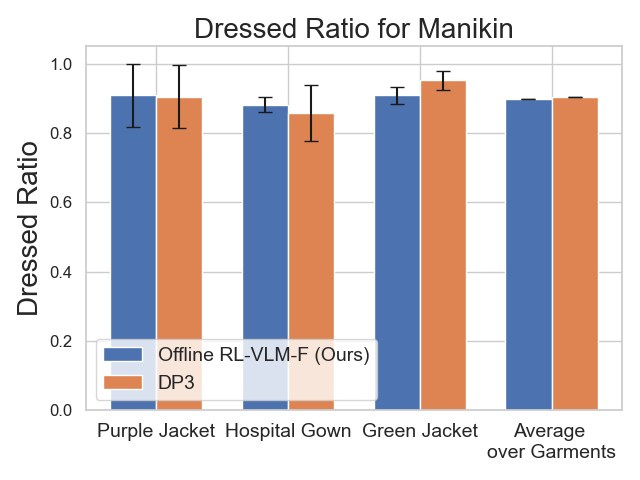}
    \vspace{-0.1in}
    \caption{Dressed ratio of our method and DP3 on the manikin arm. Both methods achieve similarly high performance, as the manikin arm better represents a real person's arm and holds an easier pose for dressing.}
    \label{fig:real-world-manikin}
    \vspace{-0.2in}
\end{figure}

\begin{figure*}[t]
    \centering
    \includegraphics[width=\textwidth]{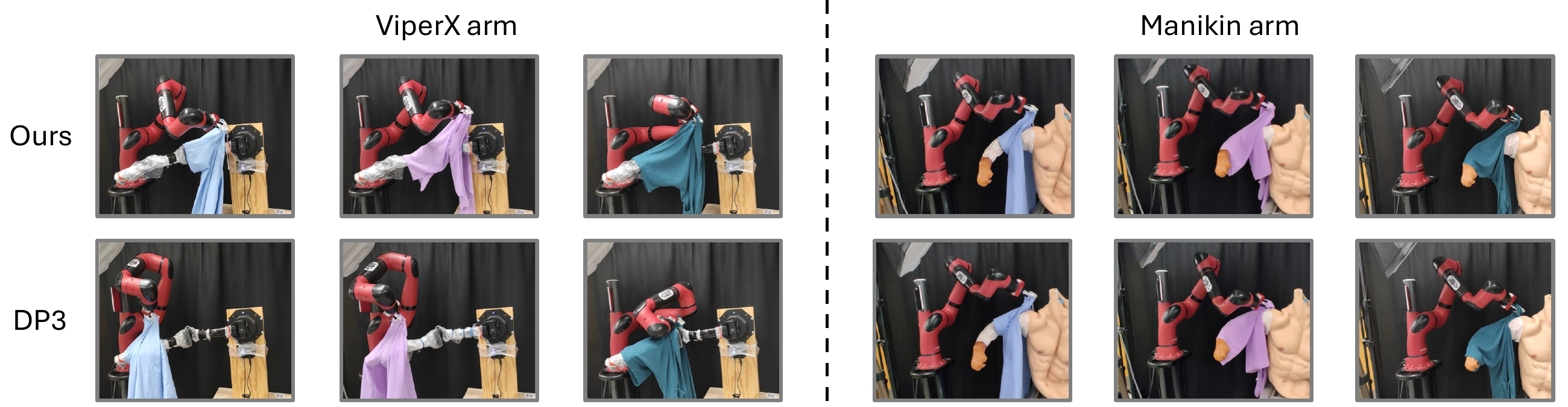}
    \caption{Left: the final dressing results on the ViperX 300 S arm. Right: the final dressing results on the manikin arm. Top: \method{} (our method). Bottom: DP3. As shown, our method achieves a better dressing result than DP3. }
    \label{fig:real_world_visual}
    \vspace{-0.2in}
\end{figure*}

\subsection{Real-World Robot-Assisted Dressing}
\subsubsection{Setup}
We also test our method in a real-world robot-assisted dressing task, where the goal is for a Sawyer robot to dress one sleeve of a garment to a person's shoulder. As in prior work~\cite{wang2023one, sun2024force}, we assume the garment is already grasped and that the person remains static during the dressing process. 
To ensure a controlled comparison between methods, we use both a manikin and a robot arm (ViperX 300 S) as a substitute for a human limb being dressed. 
The rightmost two images in Fig.~\ref{fig:tasks} illustrate the real-world setup. 

The offline dataset we use is collected from a prior human study in Wang et al.~\cite{wang2023one}. The policy used to collect the dataset is trained using reinforcement learning and policy distillation in simulation and then transferred to the real world. Due to sim2real gaps, the policy is not optimal in the dressing task, and thus, the offline dataset is not optimal and contains failure trajectories. There are in total 485 trajectories, which correspond to 26158 transitions in this offline dataset. As garments do not have a known compact state representation, the logged observation in the dataset is the segmented point cloud of the scene, which contains the garment point cloud, the arm point cloud, and a point that corresponds to the robot end-effector, which we use to train the reward and the policy. We randomly sampled 4000 image pairs to query the VLM for preference labels. The VLM we use for this real-world dressing task is GPT-4o. 

We compare to a state-of-the-art behavior cloning baseline that takes point cloud as input: 3D diffusion policy (DP3)~\cite{ze20243d}. The evaluation metric is the arm dressed ratio, which is the ratio between the dressed distance on the arm and the total length of the arm. 
We test both methods in two scenarios: the first is dressing the manikin, which holds a ``L'' shape arm pose. The second is dressing the ViperX 300 S arm in a pose akin to a human stretching their forearm away from their torso. We note that neither of these two arms is present in the training set, which only includes data collected from real people. 
The shape and morphology of the manikin arm are similar to those of real people, and the shape and morphology of the ViperX 300 S arm are rather distinct from arms of real people.
Each dressing trial stops after 60 steps, or when the sensed force on the Sawyer end-effector is larger than 7 Newtons. 
We test each method with 3 different garments, as shown in Fig.~\ref{fig:garment}. For both the manikin and the ViperX 300 S arm, we run each method on each garment 5 times and report the mean and standard deviation of the dressed ratios.

\subsubsection{Real-world results}
The results are presented in Fig.~\ref{fig:real-world-viper} and Fig.~\ref{fig:real-world-manikin}. As shown, both methods perform well when dressing the manikin, achieving an average dressed ratio of 0.9.  More interestingly, when dressing the ViperX 300 S arm, our method achieves a much higher average dressed ratio of 0.83 compared to 0.32 achieved by DP3. Fig.~\ref{fig:real_world_visual} shows the final dressed states of both methods in these two test cases.
When testing DP3 with the ViperX arm, we notice that even though the elbow is extended outwards from the body, the DP3 policy always tries to move forward instead of following the direction of the forearm. As a result, it often gets the garment stuck on the forearm and cannot progress with dressing (see Fig.~\ref{fig:real_world_visual}). 
This could be because the shape and morphology of the ViperX 300 S arm is rather distinct from arms of real people. This might explain why DP3 (behavior cloning) performs poorly in this case, as the offline dataset is not optimal, and behavior cloning (DP3) is known to generalize poor towards out-of-distribution cases. 
Instead of just imitating the trajectories in the offline dataset that include failure trajectories, our method performs offline RL and optimizes the true task objective of dressing the arm, leading to better performance. 
Videos of the dressing trials can be found on our \href{https://offline-rlvlmf.github.io/}{project website}. 
This demonstrates that our system can generate effective policies with sub-optimal, unlabeled dataset directly in the real world.  

\subsubsection{VLM Limitations and Failure Modes} We observed multiple failure modes of our approach. First, the VLM can misinterpret visual context due to camera angles—for example, it occasionally fails to detect the robot grasping the garment and incorrectly assumes the person is already dressed. Second, in early time steps of the dressing trajectories, the visual cues for progress (such as changes in cloth height or deformation) can be too subtle, leading the VLM to produce inaccurate preference labels. Third, we observed hallucinations in certain cases—e.g., the VLM mistakenly identified a hanger in the scene, which was not present. Lastly, when the two queried images are visually similar—especially when sampled from adjacent time steps—the VLM produces noisy preferences, and the query is not informative. To mitigate this, we only compare frames that are beyond a threshold number of timesteps apart.  
This ensures greater visual distinction between comparisons and improves label quality.

\section{CONCLUSIONS}
In this paper, we propose \method{}, a new system that enables automatic reward labeling and policy learning from unlabeled, sub-optimal offline datasets. We build on prior work, RL-VLM-F, to automatically generate reward labels for offline datasets using preference feedback from a vision-language model and a text description of the task. 
 Our method then learns a policy using offline RL with the reward-labeled dataset. 
 We test our system in a complex real-world robot-assisted dressing task, where we first learn a reward function using a vision-language model 
on a sub-optimal offline dataset, and then we 
use the learned reward to employ Implicit Q Learning to develop an effective dressing policy. Our method also performs well in simulation tasks involving 
the manipulation of rigid and deformable objects, outperforming baselines such as behavior cloning and inverse RL. In summary, we propose a new system that enables automatic reward labeling and policy learning from unlabeled, sub-optimal offline datasets. 

\section{Acknowledgments}
This material is based upon work supported by the National Science Foundation under Grant No. IIS-2046491 and the National Institute of Standards and Technology (NIST) under Grant No. 70NANB24H314. Any opinions, findings, and conclusions or recommendations expressed in this material are those of the authors and do not necessarily reflect the views of the National Science Foundation or NIST.
\addtolength{\textheight}{-1.5cm}   







\bibliographystyle{ieeetr}
\bibliography{main}

\end{document}